\documentclass[11pt]{article}

\usepackage[final]{acl}

\usepackage{times}
\usepackage{latexsym}

\usepackage[T1]{fontenc}

\usepackage[utf8]{inputenc}

\usepackage{inconsolata}

\usepackage{graphicx}

\usepackage{times}
\usepackage{latexsym}
\usepackage[T1]{fontenc}
\usepackage[utf8]{inputenc}
\usepackage{microtype}
\usepackage{inconsolata}

\usepackage{graphicx}
\usepackage{booktabs}
\usepackage{multirow}
\usepackage{subcaption}
\usepackage{float}
\usepackage{makecell}

\usepackage{amsmath}
\usepackage{amssymb}
\usepackage{mathtools}
\usepackage{amsthm}

\usepackage{algorithm}
\usepackage{algorithmic}

\usepackage[most]{tcolorbox}
\newtcolorbox{promptbox}[1]{enhanced, breakable, colback=gray!4,
  colframe=gray!55, coltitle=black, fonttitle=\bfseries\footnotesize,
  fontupper=\footnotesize, title={#1}, boxrule=0.4pt, arc=2pt,
  left=5pt, right=5pt, top=3pt, bottom=3pt}

\usepackage{pifont} 
\usepackage{xcolor}
\newcommand{\cmark}{\ding{51}}
\newcommand{\xmark}{\ding{55}}

\newcommand{\ql}[1]{}
\newcommand{\qlcut}[1]{}

\usepackage[capitalize,noabbrev]{cleveref}

\theoremstyle{plain}

\theoremstyle{definition}

\theoremstyle{remark}

\title{Learning What to Retain: Gated-Memory Routing for Efficient Collaboration in Multi-Agent LLM Systems}

\author{
  Rakibul Hasan Rajib \and
  Mengxin Zheng \and
  Qian Lou \\
  University of Central Florida
}

\begin{document}
\maketitle

\begin{abstract}
Large language model (LLM)--based multi-agent systems tackle complex reasoning
by orchestrating how multiple agents are configured and how they collaborate. A
central challenge is to adapt orchestration to the evolving collaboration state.
Routing from the query alone cannot adapt to intermediate progress or errors, which hurts accuracy.
Routing from the complete execution history supplies this missing context, but
forces later decisions to process every prior step, including redundant or
low-utility ones. This creates an \emph{execution-history overload} that inflates
cost. Effective orchestration
instead requires a compact state that captures useful progress without accumulating
redundant context. We propose \textbf{Gated-Memory Routing}, which conditions
each decision on the query and a learned execution memory. A learned \emph{Memory
Write Gate} commits only non-redundant reasoning steps, and a
learned \emph{Retrieval Gate} supplies each agent a compact, relevant subset, so
every decision conditions on a clean, informative state. At each step, the system
selects the next role and backbone from this memory, while an
\emph{Adaptive Halting Controller} stops execution once the memory contains sufficient evidence for answering. Across five
reasoning and code-generation benchmarks, our framework is both effective and
efficient: it attains the best average accuracy, exceeding the strongest baseline
by $2.44$ points, while reducing HumanEval inference cost by $31.9\%$ relative to
that baseline. Code is available at \url{https://github.com/rajibrhasan/gated-memory-routing}.
\end{abstract}

\section{Introduction}

\begin{figure}[t]
    \centering

    \begin{subfigure}{\columnwidth}
        \centering
        \includegraphics[width=\columnwidth]{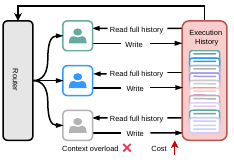}
        \caption{Full-History Routing}
        \label{fig:full-history}
    \end{subfigure}

    \begin{subfigure}{\columnwidth}
        \centering
        \includegraphics[width=\columnwidth]{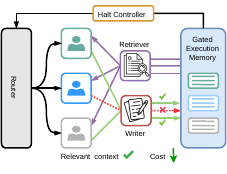}
        \caption{Gated-Memory Routing}
        \label{fig:gated-memory}
    \end{subfigure}

  \caption{
    Full-history routing feeds each decision the entire trajectory, raising context length and cost.
   Gated-memory routing instead maintains a selective memory through write and retrieval gates and halts when further collaboration is unlikely to add value, keeping context relevant and cost low.
    }
    \label{fig:routing-paradigms}
\end{figure}

Large language model (LLM)--based multi-agent systems have become a powerful
paradigm for complex reasoning: by decomposing a problem across agents that
specialize, critique, or extend one another's reasoning, they achieve strong
performance on reasoning and code-generation
benchmarks~\cite{chen2024agentverse, liang2024encouraging, zhang-etal-2024-exploring, qian2025scaling}.
Their effectiveness depends not only on the underlying LLMs but on how execution
is \emph{orchestrated}: how many agents to invoke, which roles to assign, which
backbone powers each agent, and what each agent may see. These decisions
constitute the \emph{routing} problem, and how well they are made determines
whether collaboration yields genuine reasoning gains or redundant computation.

Early multi-agent systems address routing through fixed, hand-engineered
designs, instantiating predetermined agents with manually specified roles and
static collaboration patterns that remain constant across
queries~\cite{hong2024metagpt, qian2024chatdev, li2023camel, du2024improving, wu2024autogen}.
To relax this rigidity, later work makes the interaction structure adaptive by
optimizing or pruning the communication
graph~\cite{zhuge2024gptswarm, zhang2025cut, wang2025agentdropout, liu2023dynamic},
and end-to-end routers configure agent count, roles, and backbones directly from
the query~\cite{yue2025masrouter}. These \emph{query-only} routers commit
routing decisions before any agent produces reasoning, so they cannot respond to
partial progress, errors, or emerging gaps in the solution. A recent alternative
conditions routing on the unfolding execution, sequencing agents over the
evolving task state~\cite{dang2025evolving}. When that evolving state is carried
forward with little filtering, however, each decision must reason over an
undifferentiated, ever-growing context that mixes useful evidence with redundant
or erroneous reasoning; this \emph{full-history routing} exposes downstream
agents to low-value context and inflates inference cost.

These two approaches have opposite limitations: conditioning on the query alone observes too little, committing before execution begins, which hurts accuracy, while conditioning on the full execution history (Figure~\ref{fig:routing-paradigms}\subref*{fig:full-history}) observes too much, forcing each decision to process an unfiltered and growing trajectory, which inflates cost. A router should instead learn what to retain, what to surface, and how to act on the resulting state. This motivates conditioning routing on a \emph{structured, selective memory} (Figure~\ref{fig:routing-paradigms}\subref*{fig:gated-memory}): the system dynamically routes based on a gated memory, exposes only step-relevant context to downstream agents, and commits only the reasoning steps worth preserving. These decisions are tightly coupled: low-quality writes pollute later retrieval, overly broad retrieval distracts downstream agents, and fixed-depth execution either stops prematurely or continues spending computation after additional reasoning is unlikely to help.

Based on this view, we introduce \textbf{Gated-Memory Routing}, a framework in which a learned, gated execution memory serves as the shared state that every routing decision conditions on. A single jointly trained router acts on it through complementary components: a \emph{History-Aware Role Allocator} and an \emph{LLM Router} choose the agent and its backbone, a \emph{Retrieval Gate} surfaces a compact, step-relevant subset of memory, a \emph{Memory Write Gate} commits only high-utility, non-redundant reasoning steps, and an \emph{Adaptive Halting Controller} decides when the gated state is sufficient to stop. Because routing is driven by this gated memory rather than the raw history, the interaction structure emerges dynamically during execution.


\paragraph{Contributions.}
\begin{itemize}
    \item \textbf{Routing over a learned gated memory.} We recast multi-agent
    orchestration as routing over a learned, selective execution memory: each
    step conditions on a filtered task-relevant state rather than the query alone (query-only)
    or the raw execution history (full-history).

    \item \textbf{Memory curation and budget-aware halting.} Learned write and
    retrieval gates control what memory stores and surfaces, keeping the state compact and high-signal,
    while a budget-aware halting policy consumes that clean state to control
    reasoning depth and thus cost, all trained end-to-end with role and backbone
    routing under a group-relative, cost-aware objective.

    \item \textbf{Empirical effectiveness and efficiency.} Across MATH, GSM-Hard,
    MBPP, HumanEval, and MMLU-Pro, Gated-Memory Routing attains the best average
    accuracy, exceeding the strongest baseline by $2.44$ points while cutting
    HumanEval inference cost by $31.9\%$ relative to that baseline.
\end{itemize}

\section{Related Work}

\paragraph{Fixed and Role-Based Multi-Agent Systems.}
Representative LLM-based multi-agent systems design collaboration through
predefined roles and static interaction graphs. MetaGPT~\cite{hong2024metagpt}
enforces role-based workflows through Standard Operating Procedures, while
ChatDev~\cite{qian2024chatdev} models software development as a fixed linear
chain. These systems demonstrate the value of specialization, but their
communication patterns are fixed before execution begins and cannot adapt the
participating agents, context, or depth as execution unfolds.

\paragraph{Adaptive Topology and Graph Optimization.}
Several methods make workflows more flexible by adapting interaction structures or optimizing communication graphs. GPTSwarm~\cite{zhuge2024gptswarm} optimizes
orchestration through learnable prompting, and AFlow~\cite{zhang2025aflow}
searches over executable workflows. AgentPrune~\cite{zhang2025cut} and
AgentDropout~\cite{wang2025agentdropout} induce sparsity by pruning
connections or agents from predefined templates. DyLAN~\cite{liu2023dynamic}
goes further by pruning low-contribution agents during inference using peer
evaluation and consensus, and OMAC~\cite{li2026omac} jointly optimizes agent
functionality and collaboration structure. These methods improve flexibility by
optimizing which agents participate and how they are connected, but they do not
learn which intermediate information subsequent orchestration decisions should
condition on; DyLAN, for example, adapts by removing agents while still
forwarding the unfiltered outputs of the remaining ones.

\paragraph{Query-Level LLM and MAS Routing.}
LLM routing optimizes the trade-off between model cost and capability. RouteLLM~\cite{ong2025routellm}, RouterDC~\cite{chen2024routerdc}, R2-Router~\cite{xue2026rrouter}, HW-Router~\cite{kabir2026hwrouter}, Securerouter~\cite{zhang2026securerouter} and
FrugalGPT~\cite{chen2024frugalgpt} route each query to a cost-effective model.
In multi-agent systems, MASRouter~\cite{yue2025masrouter} trains an RL
policy to configure execution components, including agent count, role assignment,
and backbone selection, from the input query. These methods make routing more
cost-aware and task-adaptive, but their decisions are primarily conditioned on
the query rather than on the unfolding execution trajectory. They therefore
cannot revise orchestration based on the quality, novelty, or redundancy of
intermediate trajectory elements.

\paragraph{Routing over the Execution History.}
Recent work addresses the query-only routing limitation by conditioning orchestration on
the unfolding execution state. Evolving Orchestration~\cite{dang2025evolving}
uses RL to train a centralized orchestrator that selects agents from the
evolving execution state and terminates through a designated stopping action.
This makes routing responsive to execution, but the state retains the
accumulated trajectory with limited filtering, so intermediate elements are
carried forward to later decisions. Such full-history routing can lead to
execution-history overload as execution deepens, forcing later decisions to
process redundant, noisy, or low-utility intermediate content. Our framework
instead routes through a gated execution memory, conditioning each decision on a
filtered state rather than the unfiltered history, and outperforms Evolving
Orchestration by $2.44$ average points across the benchmarks in
Table~\ref{tab:main_results}. Our adaptive halting also
relates to adaptive computation~\cite{graves2016act} and early-exit inference,
which learn when to stop computing within a single model; in contrast, it halts
multi-agent collaboration based on the gated memory state.

\paragraph{Memory-Optimized Agent Architectures.}
Memory management is a core component of agent systems.
MemGPT~\cite{packer2023memgpt} pages information in and out of context through a
virtual memory hierarchy, AIOS~\cite{mei2025aios} manages memory as a shared
OS-level resource for agents, and systems such as A-mem~\cite{xu2025amem} and
MM~\cite{hatalis2023memory} organize memory through categorization and retrieval.
Memory-R1~\cite{yan2025memoryr1} uses reinforcement learning to train a single
agent's policies for writing to and retrieving from a long-term memory over an
external store. These systems primarily treat memory as a storage and retrieval substrate,
optimizing what information should be preserved or recalled under a context
budget. In contrast, Gated-Memory Routing treats the within-query execution memory
of a multi-agent trajectory as an active control signal:
write and retrieval decisions are trained jointly with role allocation, backbone
routing, and halting under the task-level reward, so memory not only supplies
context but also determines who acts next, what they read, and when
collaboration stops.

\section{Methodology}
\label{sec:method}
\begin{figure*}[t]
    \centering
    \includegraphics[width=.9\textwidth]{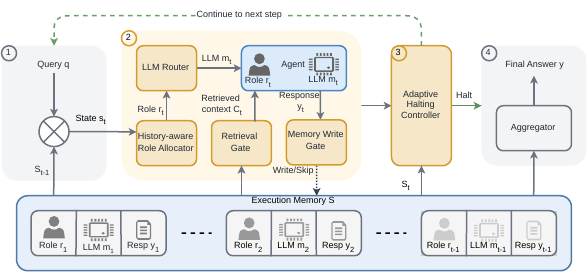}
    \caption{
    \textbf{Gated-Memory Routing framework.}
    At each step $t$, the system constructs a state $s_t=(q,S_{t-1})$ from the query
    and current execution memory. A role allocator and LLM router configure the next
    agent, a retrieval gate supplies relevant memory context, and a write gate
    decides whether the new execution record enters memory. The halting controller
    either continues the loop or passes the stored records to an Aggregator LLM for
    the final answer.
    }
    \label{fig:framework}
\end{figure*}

\subsection{Problem Definition}

We formalize gated-memory routing as a sequential decision process over a
set of agent specializations $\mathcal{R}$, a pool of heterogeneous LLM backbones
$\mathcal{M}$, and an execution memory $S_t$ maintained within a single query
execution. At step $t$, the observable state is $s_t=(q,S_{t-1})$, where
$S_{t-1}$ holds the retained execution records from prior steps. The system
selects a specialization $r_t \in \mathcal{R}$ and a backbone
$m_t \in \mathcal{M}$, retrieves a context subset $C_t \subseteq S_{t-1}$,
generates a reasoning step $y_t$, and decides whether to commit the record
$e_t=(r_t,m_t,y_t)$, so memory evolves as a gated subset of the trajectory,
$S_t \subseteq \{e_i\}_{i=1}^{t}$. The process terminates at a realized depth
$L \in \{1,\dots,\phi\}$ when the halting policy fires or the maximum depth
$\phi$ is reached; because halting is evaluated after each step, at least one
agent always acts. An aggregator then produces the final answer from the terminal
state $S_L$. Throughout, boldface denotes the frozen sentence embedding of a
quantity (e.g., $\mathbf{q}=\operatorname{enc}(q)$).

Writing $\tau_{<t}=\{e_i\}_{i=1}^{t-1}$ for the full raw trajectory before step
$t$, routing strategies differ in the state each decision sees:
\emph{query-only} routing conditions on $q$ alone, \emph{full-history} routing on
$q$ and the entire $\tau_{<t}$, and \emph{gated-memory} routing on
$s_t=(q,S_{t-1})$, pairing the query with a learned selective memory derived from
$\tau_{<t}$ that keeps only selected records.

\subsection{Framework Overview}

Figure~\ref{fig:framework} illustrates Gated-Memory Routing. Its central object is the gated execution memory $S_t$ that every decision conditions on. A single jointly trained router acts over it through complementary components: a \emph{Memory Write Gate} (Section~\ref{sec:write_filter}) and \emph{Retrieval Gate} (Section~\ref{sec:context_retriever}) keep the memory high-signal, so the \emph{History-Aware Role Allocator} (Section~\ref{sec:role_selector}) and \emph{LLM Router} (Section~\ref{sec:llm_router}) choose from a gated execution state, while the \emph{Adaptive Halting Controller} (Section~\ref{sec:halting}) reads that state to decide when to stop, making depth, and thus cost, adaptive. Because the memory stays compact and trustworthy, it can halt early without the overload full-history routing incurs. The trajectory is trained end-to-end with a single reward trading answer quality against cost (Section~\ref{sec:optimization}); encoders and representations appear in Appendix~\ref{app:representations}.

\subsection{History-Aware Role Allocator}
\label{sec:role_selector}

Role allocation chooses which agent specialization acts at each step. Rather
than committing from the query alone or reacting to the full raw trajectory, our
allocator conditions on the current state $s_t=(q,S_{t-1})$, so the next
specialization complements the retained solution state rather than every
intermediate step produced.

Because specializations such as a critic, verifier, or debugger carry semantic
structure rather than being interchangeable IDs, we represent each role through
its textual description rather than a categorical label. The description is
encoded and projected through a
Variational Autoencoder (VAE)~\cite{kingma2013auto} into a continuous latent role
embedding $\mathbf{r}_i$. A state encoder maps $s_t=(q,S_{t-1})$ to a context
vector $\mathbf{c}_t$, and the allocator defines a stochastic policy over the
available roles $\mathcal{R}$:
\[
\pi_r(r_i \mid s_t) =
\frac{\exp(\mathbf{c}_t^\top \mathbf{r}_i)}
{\sum_{r_j \in \mathcal{R}} \exp(\mathbf{c}_t^\top \mathbf{r}_j)}.
\]
This policy selects the specialization whose latent description best matches
the query and current memory state, supporting transfer across related roles and
extension to newly described specializations without changing the routing
interface.

\subsection{LLM Router}
\label{sec:llm_router}

Backbone selection matches the current reasoning need to a model's capability
and cost. Because this need depends on the selected role and the progress
already in memory (a difficult step may warrant a stronger model, while memory
may already supply enough guidance for a cheaper one), we model LLM routing as a
role-conditioned policy over the current state $s_t$ rather than a query-level
choice.

The router combines the state context $\mathbf{c}_t$ with the selected role
representation via a learned projection to form a role-conditioned context
$\mathbf{u}_t$. Each candidate LLM is represented by a continuous latent
capability vector $\mathbf{m}_i$, derived from its natural-language description
using the same description-encoding scheme as the role allocator, rather than by
a fixed index. The router samples the backbone according to
\[
\pi_m(m_i \mid s_t,r_t) =
\frac{\exp(\mathbf{u}_t^\top \mathbf{m}_i)}
{\sum_{m_j \in \mathcal{M}} \exp(\mathbf{u}_t^\top \mathbf{m}_j)}.
\]
This lets the system route each step to a model whose described capabilities
match the current role and memory state.

\subsection{Retrieval Gate}
\label{sec:context_retriever}

Even a selective memory should not be exposed wholesale to every agent: many
retained records belong to other subproblems or roles irrelevant to the agent
about to act, so passing all of them inflates token cost and buries the entries
that matter. A fixed top-$k$ retriever is equally rigid, since some steps need
broad context while others need one record or none. The \emph{Retrieval Gate}
therefore treats context construction as a step-adaptive selection, deciding
independently for each record whether to surface it.

Each stored record is summarized by an embedding $\mathbf{v}_j$ that fuses its
role latent, backbone latent, and a projection of its response, and the current
step forms a retrieval query $\mathbf{p}_t$ by the analogous fusion of the query
with the role and backbone just selected. The gate therefore asks which prior
records a step of this configuration should read: for record $j$, the retrieval
logit and binary decision are
\[
\ell_{t,j} = \mathbf{p}_t^\top \mathbf{v}_j,
\qquad
z_{t,j} \sim \mathrm{Bernoulli}(\sigma(\ell_{t,j})),
\]
and the retrieved context is
\[
C_t = \{e_j \in S_{t-1}: z_{t,j}=1\}.
\]
Because the draws are independent, $C_t$ varies in size, contracting toward
empty when memory is largely irrelevant and expanding when several records bear
on the step. Independence keeps each decision local; cross-record redundancy is
instead suppressed at write time (Section~\ref{sec:write_filter}). The gate has
no cost term of its own: retrieving more records lengthens the agent prompt and
is paid for only through the trajectory reward (Section~\ref{sec:optimization}).
\subsection{Memory Write Gate}
\label{sec:write_filter}

What the system writes to memory determines the state on which all later
decisions are based. If every intermediate element is appended, gated memory
collapses back into full-history routing, and redundant records crowd memory and
dilute retrieval. The \emph{Memory Write Gate} therefore treats memory
construction as a selective decision rather than a default append: a new record
should enter memory only when it is both relevant to the task and novel with
respect to the records already stored.

We score this relevance--novelty trade-off with a Maximal Marginal Relevance
criterion~\cite{carbonell1998use}, but replace its fixed components with learned,
stochastic ones. Let $\mathbf{c}_t$ denote the current context state
(Section~\ref{sec:role_selector}), $\mathbf{y}_t$ the embedding of the new
reasoning step, and $\{\mathbf{y}_j\}$ the embeddings of stored steps, and let
$\mathrm{sim}(\cdot,\cdot)$ be a cosine similarity taken in a learned projection
of these frozen embeddings, so the relevance and novelty geometry is trained
rather than prescribed. The write score is
\[
\omega_t =
\lambda \, \mathrm{sim}(\mathbf{y}_t, \mathbf{c}_t)
-
(1-\lambda)\max_{j:e_j\in S_{t-1}} \mathrm{sim}(\mathbf{y}_t, \mathbf{y}_j),
\]
with a learned coefficient $\lambda \in (0,1)$ balancing relevance against
redundancy with stored records; the novelty term is dropped when memory is
empty, so the first record is admitted on relevance alone. Rather than greedily
keeping the top-scoring record as in deterministic MMR, we sample the write
decision $w_t \sim \mathrm{Bernoulli}(\sigma(\omega_t))$, keeping memory
construction trainable under the same policy objective as the routing decisions. The novelty criterion controls cross-record redundancy, allowing the retrieval gate (Section~\ref{sec:context_retriever}) to score records independently. As a result, memory remains a compact representation of execution progress rather than accumulating into a transcript.

\subsection{Adaptive Halting Controller}
\label{sec:halting}

Reasoning depth is a major driver of cost, yet a depth fixed from the query
commits this budget before the system observes how execution unfolds. We
therefore make depth a consequence of execution: after each step the
\emph{Adaptive Halting Controller} decides whether the memory holds sufficient
evidence for aggregation or another agent should act.

Whether recent steps still add evidence is a trend no single-step snapshot
captures, so the controller carries a recurrent state
$\mathbf{h}^{\mathrm{halt}}_t = \mathrm{GRU}(\mathbf{h}^{\mathrm{halt}}_{t-1},
\mathrm{enc}(S_t))$ that integrates a pooled summary of the post-write memory
across steps, kept separate from the query-attended routing context
$\mathbf{c}_t$. It then samples a halt action
\[
h_t \sim \mathrm{Bernoulli}\!\left(\sigma\!\left(\mathrm{MLP}(\mathbf{h}^{\mathrm{halt}}_t)\right)\right).
\]
Because the decision is evaluated only after the step's agent has executed, at
least one agent always runs. If $h_t=1$ the stored records pass to the
Aggregator LLM; otherwise execution continues to the maximum depth $\phi$. The
halt action is optimized under the same trajectory-level reward as the other
decisions (Section~\ref{sec:optimization}), so reasoning depth co-adapts with
routing.

\subsection{Optimization}
\label{sec:optimization}

The routing stack couples several discrete stochastic decisions (role, backbone,
retrieval, write, and halt), and their number varies with the halt-determined
depth, so we optimize the trajectory with policy gradients rather than
backpropagating through sampled actions. Because queries differ widely in
difficulty, a single trajectory baseline is noisy; we instead train all policies
jointly with a group-relative advantage in the spirit of
GRPO~\cite{shao2024deepseekmath}, using $G$ rollouts of each query as a per-query
baseline without a learned critic.

For each query we sample $G$ trajectories that differ only in their sampled
decisions. Trajectory $\tau_i$ has utility $u_i = R_i - \lambda_c\,\mathrm{Cost}(\tau_i)$,
where $R_i\in\{0,1\}$ is task success, $\mathrm{Cost}(\tau_i)$ sums the per-step
backbone costs, and $\lambda_c$ sets the accuracy--cost trade-off
(Appendix~\ref{app:cost}). Centering each utility
within its group yields the advantage
\[
A_i = u_i - \frac{1}{G}\sum_{i' \in \mathcal{G}(i)} u_{i'},
\]
a per-query baseline that cancels difficulty without a critic. We deliberately
drop the usual within-group standard-deviation normalization: when rollouts for
a query share near-identical rewards, the standard deviation approaches zero and
inflates the advantage from negligible cost differences, whereas the centered
advantage remains well-scaled. The same advantage weights every sampled action. With $\log\pi_i$ the summed
log-probability of all role, backbone, retrieval, write, and halt actions in
$\tau_i$, the objective is
\[
\mathcal{L} =
-\mathrm{mean}_i\!\left[\log\pi_i\,\mathrm{sg}(A_i)\right]
+ \alpha\,\mathcal{L}_{\mathrm{VAE}}
- c_H\,\bar{H}(\pi),
\]
where $\mathrm{sg}$ is stop-gradient, $\mathcal{L}_{\mathrm{VAE}}$ regularizes
the role and backbone encoders (Appendix~\ref{app:representations}), and
$\bar{H}(\pi)$ is the mean per-step entropy of the routing and halt policies. The
entropy term discourages premature deterministic behavior;
Appendix~\ref{app:gate_behavior} verifies empirically that the trained gates do
not collapse to trivial always-write, never-write, retrieve-all, or
retrieve-none solutions. The
LLM backbones and sentence encoder are frozen, so the router is the only trained
component, updated jointly by a single optimizer.

\section{Experimental Setup}
\label{sec:setup}

\paragraph{Datasets and Benchmarks.}
We evaluate on five benchmarks spanning mathematical reasoning, program synthesis, and knowledge-intensive question answering: GSM-Hard~\cite{gao2023pal}, MATH~\cite{hendrycks2021math}, HumanEval~\cite{chen2021evaluating}, MBPP~\cite{austin2021program}, and MMLU-Pro~\cite{wang2024mmlu}.

\paragraph{Baselines.}
Our baselines span single-model inference; single-agent reasoning (CoT~\cite{wei2022chain}, Complex-CoT~\cite{fu2023complexitybased}); fixed-topology multi-agent collaboration (MacNet-Chain, -Tree, and -Complete~\cite{qian2025scaling}); adaptive workflow optimization (AFlow~\cite{zhang2025aflow}); evolving orchestration (Puppeteer~\cite{dang2025evolving}); and query-only routing (MASRouter~\cite{yue2025masrouter}).

\paragraph{LLM Backbones.}
We route over five open-weight LLMs: llama-3.2-3B, llama-3.1-8B~\cite{grattafiori2024llama}, mistral-nemo-12B~\cite{mistral2024nemo}, qwen-2.5-14B, and qwen-2.5-32B~\cite{yang2025qwen3}. This 3B--32B heterogeneous pool allows the router to trade off capability and inference cost. To quantify the accuracy--cost trade-off, we assign each backbone a price proportional to its parameter count, a hardware-independent proxy for per-token inference compute (Appendix~\ref{app:cost}). Backbone capability descriptions are listed in Appendix~\ref{app:llm_profiles}.

\subsection{Implementation Details}
\label{sec:implementation_details}

We train with Adam optimizer using learning rate $0.01$, GRPO group size $G=6$, and $16$ queries per batch. We set maximum depth $\phi=6$, cost penalty $\lambda_c \in \{10, 20, 50\}$, entropy coefficient $c_H=0.01$, and VAE regularizer weight $\alpha=0.001$. For the fixed-topology multi-agent baselines, we set the number of agents equal to our maximum depth $\phi=6$, so every multi-agent method operates under the same agent budget. We use the same $26$ heterogeneous role profiles as MASRouter~\cite{yue2025masrouter}, spanning programming agents with compiler access to research-oriented agents with external knowledge sources; full descriptions are in Appendix~\ref{app:roles}. For final aggregation, we use the LLM backbone selected most frequently across the query's reasoning steps. The aggregation prompt is provided in Appendix~\ref{app:aggregator_prompt}.

\begin{table*}[t]
\centering
\scriptsize
\setlength{\tabcolsep}{3pt}
\renewcommand{\arraystretch}{1.05}
\begin{tabular*}{\textwidth}{@{\extracolsep{\fill}}llcccccccc@{}}
\toprule
Method & LLM & Mul. & Rout. & MATH & GSM-Hard & MBPP & HumanEval & MMLU-Pro & Avg. \\
\midrule
\multirow{5}{*}{Single LLM}
& llama-3.2-3B     & \xmark & \xmark & 43.27 & 25.85 & 52.20 & 62.79 & 34.00 & 43.62 \\
& llama-3.1-8B     & \xmark & \xmark & 43.75 & 39.87 & 57.60 & 69.78 & 48.25 & 51.85 \\
& mistral-nemo-12B & \xmark & \xmark & 42.07 & 30.11 & 56.00 & 68.22 & 27.75 & 44.83 \\
& qwen-2.5-14B      & \xmark & \xmark & 75.72 & 64.58 & 71.80 & 82.95 & 65.25 & 72.06 \\
& qwen-2.5-32B      & \xmark & \xmark & 78.37 & 61.52 & 79.00 & 84.37 & 66.02 & 73.86 \\
\midrule
\multirow{2}{*}{CoT~\cite{wei2022chain}}
& qwen-2.5-14B & \xmark & \xmark & 77.88 & 67.52 & 70.80 & 82.17 & 62.16 & 72.11 \\
& qwen-2.5-32B & \xmark & \xmark & 78.85 & 64.56 & 77.20 & 85.15 & 68.41 & 74.83 \\
\multirow{2}{*}{Complex-CoT~\cite{fu2023complexitybased}}
& qwen-2.5-14B & \xmark & \xmark & 75.96 & 66.76 & 70.40 & 80.62 & 63.07 & 71.36 \\
& qwen-2.5-32B & \xmark & \xmark & 77.16 & 65.32 & 78.40 & 85.47 & 68.41 & 74.95 \\
\midrule
\multirow{2}{*}{MacNet-Chain~\cite{qian2025scaling}}
& qwen-2.5-14B & \cmark & \xmark & 73.80 & 59.75 & \textbf{82.66} & \underline{86.82} & 61.48 & 72.90 \\
& qwen-2.5-32B & \cmark & \xmark & 77.64 & 61.93 & 80.73 & 83.80 & 67.95 & 74.41 \\
\multirow{2}{*}{MacNet-Tree~\cite{qian2025scaling}}
& qwen-2.5-14B & \cmark & \xmark & 76.68 & 60.32 & 79.84 & 82.95 & 60.68 & 72.09 \\
& qwen-2.5-32B & \cmark & \xmark & 77.88 & 62.31 & 80.71 & 86.02 & 67.73 & 74.93 \\
\multirow{2}{*}{MacNet-Complete~\cite{qian2025scaling}}
& qwen-2.5-14B & \cmark & \xmark & 77.16 & 59.00 & 80.65 & 84.50 & 61.48 & 72.56 \\
& qwen-2.5-32B & \cmark & \xmark & \underline{79.09} & 62.22 & \underline{81.20} & 86.05 & 67.16 & 75.14 \\
\midrule
\multirow{2}{*}{AFlow~\cite{zhang2025aflow}}
& qwen-2.5-14B & \cmark & \xmark & 63.46 & 58.24 & 70.00 & 85.27 & 64.66 & 68.33 \\
& qwen-2.5-32B & \cmark & \xmark & 77.64 & 67.05 & 76.40 & 84.50 & 68.41 & 74.80 \\
\midrule
\multirow{2}{*}{Puppeteer~\cite{dang2025evolving}}
& qwen-2.5-14B & \cmark & \xmark & 75.00 & 65.91 & 72.58 & 83.59 & 65.23 & 72.46 \\
& qwen-2.5-32B & \cmark & \xmark & \underline{79.09} & \underline{68.75} & 74.80 & 85.16 & \underline{68.64} & \underline{75.29} \\
\midrule
MASRouter~\cite{yue2025masrouter}
& LLM Pool & \cmark & \cmark & 74.31 & 66.00 & 79.20 & 85.16 & 66.70 & 74.27 \\
\midrule
\textbf{Ours}
& LLM Pool & \cmark & \cmark & \textbf{79.33} & \textbf{70.55} & 79.60 & \textbf{89.84} & \textbf{69.32} & \textbf{77.73} \\
\bottomrule
\end{tabular*}
\caption{Test accuracy (\%) across five benchmarks; MBPP and HumanEval report pass@1. Mul.\ indicates multi-agent execution, and Rout.\ indicates routing over the LLM backbone pool. Best results are in \textbf{bold}, and second-best results are \underline{underlined}. ``LLM Pool'' denotes the full five-model pool.}
\label{tab:main_results}
\end{table*}

\section{Results}

\subsection{Performance Analysis}

Table~\ref{tab:main_results} shows that Gated-Memory Routing achieves the
strongest overall performance: it attains the best average accuracy and the top
result on four of the five benchmarks, improving on the strongest baseline by
$2.44$ points on average. The gains are most pronounced on GSM-Hard and
HumanEval, where intermediate reasoning quality and selective context matter
most; the one exception is MBPP,
where a fixed multi-agent chain topology powered by qwen-2.5-14B edges ahead at
pass@1 but trails on every other benchmark and on average. On MBPP our method
is comparable to MASRouter ($79.60$ vs.\ $79.20$) and ahead of Puppeteer-32B
($74.80$), while several fixed MacNet configurations score highest. MBPP's
short, relatively uniform problems appear to benefit from a fixed pipeline built
on a strong code model, leaving less room for adaptive orchestration to help; the
benefit of gated memory is most visible where reasoning is longer and more
heterogeneous.

The comparison is sharpest against the two paradigms our method is built to
improve on. MASRouter routes from the query alone, committing every decision
before any reasoning exists, and trails by $3.46$ points. Puppeteer orchestrates
agents over the evolving execution state but does not route over the pool,
running every step on the largest backbone in the pool (qwen-2.5-32B), and still
trails by $2.44$ points. Gated-memory routing conditions each decision on a filtered state instead,
and is the most accurate of the three on every benchmark, even though it routes
over a heterogeneous pool rather than relying on the largest model alone. The
gain is therefore not a matter of raw scale but of how well roles, backbones, and
memory are coordinated as the trajectory unfolds.
Appendix~\ref{app:seed_variance} reports the mean and standard deviation over
three training seeds for our method and both routing baselines; the ranking of
the three methods is identical under every seed.

\begin{figure}[t]
    \centering
    \includegraphics[width=\columnwidth]{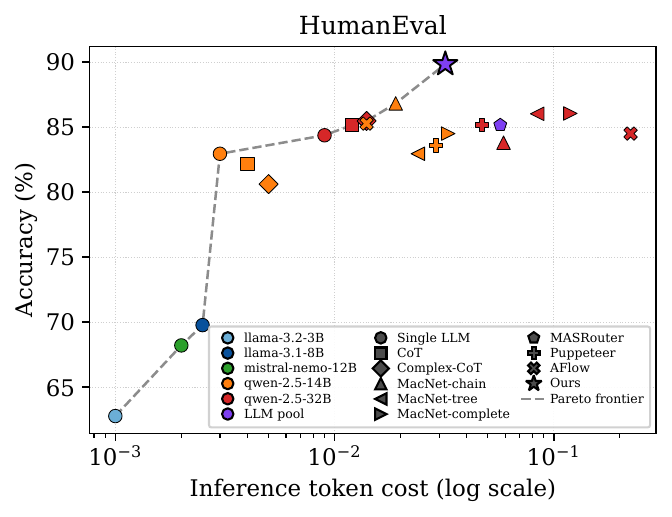}
    \caption{
    Accuracy versus inference token cost on HumanEval, with cost on a log scale.
    Our method lies on the high-accuracy end of the Pareto frontier, achieving the highest accuracy among compared methods while using substantially less inference cost than the strongest multi-agent and routing baselines. 
    }

    \vspace{-2mm}
    \label{fig:pareto}
\end{figure}

\subsection{Cost Analysis}

Figure~\ref{fig:pareto} places our method on the Pareto frontier of accuracy
against inference cost on HumanEval, where it costs $43.86\%$ less than MASRouter
and $31.9\%$ less than Puppeteer (qwen-2.5-32B) while remaining more accurate than
both, gaining on cost and accuracy at once rather than trading one for the other.
The GSM-Hard frontier in Appendix~\ref{app:cost_detail} shows the same pattern.
Appendix~\ref{app:cost} additionally reports approximate FLOPs and
batch-amortized wall-clock time per query on HumanEval and MBPP, which give the
same ordering.

These savings come from where computation is spent. Because each decision
conditions on a compact, gated memory rather than the full execution history,
the halting controller can stop once that memory carries enough evidence, so easy
queries terminate early and only the hard ones run deep. The gates contribute
indirectly: a clean memory makes halting trustworthy and spares downstream agents
the redundant context that inflates prompts under full-history routing.
Section~\ref{sec:paradigm} compares the three paradigms directly: gated-memory
routing is at least as accurate as full-history routing while costing appreciably
less, and is both more accurate and cheaper than query-only routing.

\subsection{Ablation Study}

\begin{table}[h]
\centering
\small
\setlength{\tabcolsep}{5pt}
\begin{tabular}{l|cc|cc}
\toprule
\multirow{2}{*}{Configuration} & \multicolumn{2}{c|}{GSM-Hard} & \multicolumn{2}{c}{HumanEval} \\
 & Acc. & Cost & Acc. & Cost \\
\midrule
Full system              & 70.55 & 0.587 & 89.84 & 0.032 \\
\midrule
\quad w/o role allocator & 68.37 & 0.400 & 85.94 & 0.043 \\
\quad w/o LLM router     & 56.53 & 0.555 & 71.88 & 0.026 \\
\quad w/o retrieval gate & 69.22 & 0.488 & 85.94 & 0.034 \\
\quad w/o write gate     & 66.67 & 0.621 & 84.38 & 0.037 \\
\quad w/o halting        & 70.08 & 0.840 & 86.72 & 0.086 \\
\quad w/o both memory gates & 67.99 & 0.447 & 83.59 & 0.033 \\
\bottomrule
\end{tabular}
\caption{Leave-one-out ablation. Each row disables one routing module while keeping the others active. We report accuracy (\%) and total test-set cost on GSM-Hard and HumanEval.}
\label{tab:ablation}
\end{table}

Table~\ref{tab:ablation} reports a leave-one-out ablation, each variant replacing one module with a default: \emph{w/o role allocator} uses random role selection, \emph{w/o LLM router} random backbone selection, \emph{w/o retrieval gate} exposes the full memory to each agent, \emph{w/o write gate} writes every response to memory, and \emph{w/o halting} runs every query to the maximum depth $\phi$. The effects split cleanly along the division of labor the method is built around. The two routing decisions carry accuracy: dropping the LLM router is by far the most damaging change (a $14.0$-point drop on GSM-Hard), and dropping the role allocator costs a smaller but consistent amount, so matching model capacity and specialization to the current state is what drives answer quality.

The memory gates also serve accuracy through the quality of the memory state: removing either lowers accuracy, because unfiltered context dilutes what each agent and the router condition on. Disabling both gates at once (\emph{w/o both memory gates}) lowers accuracy by $2.56$ points on GSM-Hard and $6.25$ on HumanEval, so selective memory matters even with the router and role allocat Halting is the lever that controls cost: without it, inference cost rises by over $40\%$ on GSM-Hard and more than doubles on HumanEval, while accuracy also slips, since most queries are solved before the depth limit and the extra steps add computation without improving answers. The \emph{w/o halting} variant also provides an equal-depth comparison with the fixed-topology baselines: executing all six steps, it reaches $70.08$ on GSM-Hard, above every MacNet-32B variant ($61.93$--$62.31$, Table~\ref{tab:main_results}) at lower cost (Appendix~\ref{app:cost_detail}), because each step is still routed to an appropriate backbone rather than the largest one. Halting also reduces training-time computation: disabling it raises the number of training rollout steps by about $1.8\times$ on both benchmarks (Appendix~\ref{app:halting_training}).

\begin{figure}[h]
\centering
\includegraphics[width=\columnwidth]{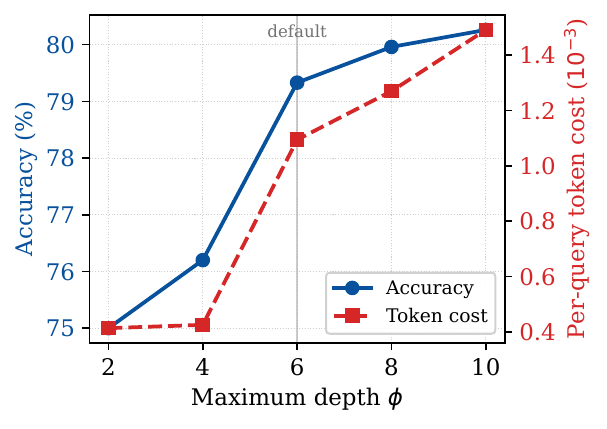}
\caption{Effect of the maximum depth $\phi$ on MATH: test accuracy (left axis) and per-query inference token cost (right axis). $\phi=6$ is the default used elsewhere in the paper.}
\label{fig:depth}
\end{figure}

\subsection{Effect of the Depth Budget}

Figure~\ref{fig:depth} varies the maximum depth $\phi$ on MATH with adaptive halting left in place. Accuracy increases with the budget but saturates: most of the gain is realized by $\phi=6$, with under a point more from $\phi=6$ to $\phi=10$ as cost keeps climbing. Small budgets truncate useful reasoning on the harder problems, while larger ones mostly add cost, since halting already stops easy queries early regardless of the ceiling. The default $\phi=6$ thus sits near the knee of the accuracy--cost curve, capturing nearly all the attainable accuracy before the budget begins buying cost more than accuracy.

\subsection{Routing-Paradigm Comparison}
\label{sec:paradigm}

We contrast Gated-Memory Routing with the two routing paradigms it is built to
improve on. Query-only routing fixes every decision from the input query before execution
begins; we instantiate it with MASRouter, the strongest query-only router among
our baselines. Under full-history routing, the per-step role and backbone allocation is
conditioned on the accumulated history, and all intermediate reasoning is
forwarded as context to the next agent. Gated-memory routing is our full system. Table~\ref{tab:paradigm} compares the three.

\begin{table}[h]
\centering
\small
\setlength{\tabcolsep}{3pt}
\begin{tabular}{lcccc}
\toprule
 & \multicolumn{2}{c}{GSM-Hard} & \multicolumn{2}{c}{HumanEval} \\
\cmidrule(lr){2-3}\cmidrule(lr){4-5}
Routing paradigm & Acc.\ (\%) & Cost & Acc.\ (\%) & Cost \\
\midrule
Query-only (MASRouter) & 66.00 & 1.012 & 85.16 & 0.057 \\
Full-history           & 70.27 & 0.985 & 89.06 & 0.068 \\
Gated-memory (ours)    & \textbf{70.55} & \textbf{0.587} & \textbf{89.84} & \textbf{0.032} \\
\bottomrule
\end{tabular}
\caption{Comparison of the three routing paradigms on GSM-Hard and HumanEval.}
\label{tab:paradigm}
\end{table}

Full-history routing conditions each decision on the entire accumulated trajectory, increasing context length and reasoning overhead. The additional context does not appear to improve
accuracy here: full-history routing is roughly on par with the curated memory
($70.27$ vs $70.55$ on GSM-Hard and $89.06$ vs $89.84$ on HumanEval), while the
gated memory attains comparable accuracy at appreciably lower cost (about $40\%$
on GSM-Hard and over $50\%$ on HumanEval). Query-only routing tends to be the
least accurate of the three. These results suggest that, in our setting,
curating the memory rather than forwarding the full history largely preserves
accuracy while reducing cost.

\section{Conclusion}
\label{sec:conclusion}

We presented \textbf{Gated-Memory Routing}, a framework for dynamic
multi-agent coordination that conditions routing decisions on a learned, gated
execution memory rather than on the query alone or the full raw history. A
Memory Write Gate and a Retrieval Gate keep this memory compact and
task-relevant, enabling role and backbone routing to operate over a filtered
execution state. An Adaptive Halting Controller uses the same memory to decide
when to stop, adapting reasoning depth and cost to each query. Across five
benchmarks, our method attains the best average accuracy, exceeding the
strongest baseline by $2.44$ points, while reducing HumanEval inference cost by
$31.9\%$ relative to that baseline. Future work will extend gated-memory routing
to open-ended generation, larger-scale agent ecosystems, and pools of more
recent backbone models, and explore richer
reward signals beyond binary task success.

\section*{Limitations}

Our evaluation uses closed-domain tasks with verifiable answers, enabling automatic reward computation; open-ended generation would require reward models, human evaluation, or other supervision. Our efficiency analysis: parameter count provides a hardware-independent compute proxy but does not capture latency, memory pressure, or batching, while FLOPs and wall-clock measurements in Appendix~\ref{app:cost} cover only two benchmarks. Finally, we evaluate open-weight Qwen2.5, Llama-3, and Mistral models; backbones can be substituted through their capability profiles and per-token costs without changing the framework.

\bibstyle{acl_natbib}
\bibliography{main}
\newpage
\section{Algorithm}
\label{app:algorithm}

We provide detailed pseudocode for Gated-Memory Routing in
Algorithm~\ref{alg:framework}.

\begin{algorithm*}[htpb]
\caption{Gated-Memory Routing}
\label{alg:framework}
\begin{algorithmic}[1]
   \STATE {\bfseries Input:} Query $q$, role set $\mathcal{R}$, model pool $\mathcal{M}$, maximum depth $\phi$
   \STATE {\bfseries Output:} Final answer $y_{\mathrm{final}}$
   \STATE Encode query embedding $\mathbf{q} \leftarrow \operatorname{enc}(q)$
   \STATE Initialize memory $S_0 \leftarrow \emptyset$, halting state $\mathbf{h}^{\mathrm{halt}}_0 \leftarrow \mathbf{0}$
   \STATE Initialize $t \leftarrow 0$, $h_0 \leftarrow 0$
   \WHILE{$h_t = 0$ {\bfseries and} $t < \phi$}
       \STATE $t \leftarrow t + 1$
       \STATE Form state $s_t \leftarrow (q,S_{t-1})$

       \STATE \textcolor{gray}{// Role and model routing}
       \STATE Sample role $r_t \sim \pi_r(\cdot \mid s_t)$
       \STATE Sample LLM backbone $m_t \sim \pi_m(\cdot \mid s_t,r_t)$

       \STATE \textcolor{gray}{// Retrieval gate}
       \STATE Form retrieval vector $\mathbf{p}_t \leftarrow
       \operatorname{Linear}([\mathbf{q} \,\|\, \mathbf{r}_{r_t} \,\|\, \mathbf{m}_{m_t}])$
       \FORALL{$e_j \in S_{t-1}$}
           \STATE Sample $z_{t,j} \sim \mathrm{Bernoulli}(\sigma(\ell_{t,j}))$
       \ENDFOR
       \STATE Retrieve context $C_t \leftarrow \{e_j \in S_{t-1}: z_{t,j}=1\}$

       \STATE \textcolor{gray}{// Agent execution}
       \STATE Generate reasoning step $y_t \leftarrow \operatorname{Agent}(q,r_t,m_t,C_t)$
       \STATE Form execution record $e_t \leftarrow (r_t,m_t,y_t)$

       \STATE \textcolor{gray}{// Memory write gate}
       \STATE Compute write score $\omega_t$ using the query, $y_t$, and stored records in $S_{t-1}$
       \STATE Sample $w_t \sim \mathrm{Bernoulli}\!\left(\sigma(\omega_t)\right)$
       \IF{$w_t = 1$}
           \STATE $S_t \leftarrow S_{t-1} \cup \{e_t\}$
       \ELSE
           \STATE $S_t \leftarrow S_{t-1}$
       \ENDIF

       \STATE \textcolor{gray}{// Adaptive halting}
       \STATE $\mathbf{h}^{\mathrm{halt}}_t \leftarrow
       \operatorname{GRU}(\mathbf{h}^{\mathrm{halt}}_{t-1}, \operatorname{enc}(S_t))$
       \STATE Sample $h_t \sim
       \mathrm{Bernoulli}\!\left(\sigma(\operatorname{MLP}(\mathbf{h}^{\mathrm{halt}}_t))\right)$
   \ENDWHILE

   \STATE $L \leftarrow t$
   \STATE Select aggregator backbone $m_{\mathrm{agg}}$ as the most frequently selected model in $\{m_i\}_{i=1}^{L}$
   \STATE $y_{\mathrm{final}} \leftarrow \operatorname{Aggregator}(q,S_L;m_{\mathrm{agg}})$
\end{algorithmic}
\end{algorithm*}

\section{Representations, Encoders, and Training Interface}
\label{app:representations}

This appendix specifies the encoders and representations underlying the routing
modules of Section~\ref{sec:method}, and how the policy is trained against the
frozen backbones.

\paragraph{Sentence encoder.}
All text is embedded by a single frozen sentence encoder $\operatorname{enc}(\cdot)$
(\texttt{all-MiniLM-L6-v2}), producing $384$-dimensional vectors; following the
convention of Section~\ref{sec:method}, $\mathbf{x}=\operatorname{enc}(x)$ for any
text $x$, so $\mathbf{q}$ is the query embedding and $\mathbf{y}_t$ the embedding
of reasoning step $y_t$. The same encoder embeds the query, the role and backbone
descriptions, and every
agent response. Its parameters are not updated during training, and response
embeddings are detached before they enter memory, so no gradient flows back
through the encoder or through the agent outputs.

\paragraph{Role and backbone latents.}
Each role profile and each backbone description is embedded once by
$\operatorname{enc}$ and mapped to a continuous latent by a variational encoder (a
small VAE with a $128$-dimensional latent), yielding the role embeddings
$\mathbf{r}_i$ and backbone embeddings $\mathbf{m}_i$ used by the role allocator
(Section~\ref{sec:role_selector}) and LLM router (Section~\ref{sec:llm_router}).
The reconstruction and KL terms of these encoders form the
$\mathcal{L}_{\mathrm{VAE}}$ regularizer in the training objective.

\paragraph{State encoder.}
The state $s_t=(q,S_{t-1})$ is encoded into the context vector $\mathbf{c}_t$
used by the routing modules. The query is linearly projected to a
$128$-dimensional vector $\hat{\mathbf{q}}$, and each retained step $j<t$ becomes
a token $\mathbf{u}_j$ formed from its role latent $\mathbf{r}_{r_j}$ and backbone
latent $\mathbf{m}_{m_j}$, additively modulated by a sigmoid gate over the
response embedding $\mathbf{y}_j$, so that response content, and not only routing
identity, informs the state. A two-layer Transformer encoder contextualizes the
tokens $\{\mathbf{u}_j\}_{j<t}$, the query attends over them to produce a pooled
history vector $\mathbf{h}_t$, and $\mathbf{c}_t = [\,\hat{\mathbf{q}} \,\|\,
\mathbf{h}_t\,]$. The execution memory is therefore encoded as an ordered set of
per-step records rather than a single averaged vector.

\paragraph{Memory-record embedding.}
The retrieval gate (Section~\ref{sec:context_retriever}) forms the query vector
$\mathbf{p}_t = \operatorname{Linear}([\mathbf{q} \,\|\, \mathbf{r}_{r_t} \,\|\,
\mathbf{m}_{m_t}])$ and, for each stored record $j$, an embedding
$\mathbf{v}_j = \operatorname{Linear}([\mathbf{r}_{r_j} \,\|\, \mathbf{m}_{m_j}
\,\|\, \tilde{\mathbf{y}}_j])$, where $\tilde{\mathbf{y}}_j$ is a learned
projection of $\mathbf{y}_j$. The retrieval logit is the scaled cosine similarity
$\ell_{t,j} = s\,\cos(\mathbf{p}_t,\mathbf{v}_j) + b$ with learnable scale $s$ and
bias $b$, and record $j$ is admitted by
$z_{t,j} \sim \mathrm{Bernoulli}(\sigma(\ell_{t,j}))$. Memory holds at most $\phi$
records, so $C_t$ ranges from empty to $\phi$ entries, with $b$ setting its
typical length.

\paragraph{Write-gate similarities.}
For the Memory Write Gate (Section~\ref{sec:write_filter}), the projection
underlying $\mathrm{sim}(\cdot,\cdot)$, the relevance--novelty weight $\lambda$,
the write threshold $\theta$, and a logit scale $\beta$ are learned jointly with
the routing policies, and the write decision is
$w_t \sim \mathrm{Bernoulli}\!\left(\sigma(\beta(\omega_t - \theta))\right)$,
which Section~\ref{sec:write_filter} abstracts as $\sigma(\omega_t)$.

\paragraph{Training interface.}
The backbones are frozen and queried as an environment: at each step the
selected model is called through an OpenAI-compatible vLLM server, and the
returned text is embedded and detached as described above. The router is
optimized purely by the score-function (policy-gradient) estimator: the
group-relative advantage $A_i$ (Section~\ref{sec:optimization}) weights the
trajectory log-probability
$\log\pi_i = \sum_{t=1}^{L}\big(\log\pi_r(r_t\mid s_t) + \log\pi_m(m_t\mid s_t,r_t)
+ \log p(z_t) + \log p(w_t) + \log p(h_t)\big)$, where $z_t$ collects the step-$t$
retrieval decisions. Gradients reach only the routing, gating, halting, and
embedding parameters; a single optimizer updates all of them jointly, and the
backbones are never differentiated through.

\section{Cost Calculation for Open-Weight Backbones}
\label{app:cost}

Because the five backbones are open-weight and served locally, they do not carry
a public per-token price. To give the cost-aware objective a hardware-independent
and monotonic cost signal, we set each model's per-token reference price
proportional to its (active) parameter count $N$ in billions: an input rate of
$0.003\,N$ and an output rate of $0.010\,N$ per million tokens, with output
weighted more heavily to reflect the higher cost of autoregressive decoding. For
dense transformers, per-token inference compute (FLOPs) grows approximately
linearly with $N$, so pricing each token by model size makes the reported cost
proportional to a size-weighted token count: a hardware-independent proxy for
relative inference compute across the pool, rather than the dollar price of any
specific API or the wall-clock cost on particular hardware. Both the training cost term of Section~\ref{sec:optimization} and the reported
test cost of a trajectory are the additive sum $\mathrm{Cost}(\tau)=\sum_t F_t$
of its per-step backbone costs, where $F_t$ is the cost of step $t$'s call (its
input and output tokens times these rates). As a relative-compute proxy it abstracts
away attention cost that scales with context length, batching, and memory
bandwidth. Table~\ref{tab:backbones} lists the backbone pool and the resulting
rates.

\begin{table}[h]
\centering
\scriptsize
\setlength{\tabcolsep}{3pt}
\renewcommand{\arraystretch}{1.05}
\begin{tabular*}{\columnwidth}{@{\extracolsep{\fill}}lcr@{}}
\toprule
Backbone & Size & Cost in/out \\
\midrule
llama-3.2-3B      & 3B  & 0.009 / 0.03 \\
llama-3.1-8B      & 8B  & 0.024 / 0.08 \\
mistral-nemo-12B           & 12B & 0.036 / 0.12 \\
qwen-2.5-14B       & 14B & 0.042 / 0.14 \\
qwen-2.5-32B       & 32B & 0.096 / 0.32 \\
\bottomrule
\end{tabular*}
\caption{Open-weight backbone pool and size-based reference rates (input $=0.003\,N$, output $=0.010\,N$ per $1$M tokens, with $N$ the parameter count in billions), shown as input/output.}
\label{tab:backbones}
\end{table}

To corroborate the size-weighted proxy with hardware-relevant measurements,
Table~\ref{tab:hw_cost} reports, for HumanEval and MBPP, an approximate FLOPs
count and the batch-amortized wall-clock time per query for our method and the
two strongest baselines. FLOPs are estimated from the per-call token records as
$2N$ FLOPs per processed token~\cite{kaplan2020scaling}, summed over all calls
with $N$ the parameter count of the model serving each call; this omits the
sequence-length-dependent attention term. Wall-clock time is the end-to-end
inference time for the test set divided by the number of queries, measured under
the same serving configuration for all methods. Our method has the lowest FLOPs
and wall-clock time on both benchmarks, and the ordering matches the proxy.

\begin{table}[h]
\centering
\scriptsize
\setlength{\tabcolsep}{3pt}
\renewcommand{\arraystretch}{1.05}
\begin{tabular*}{\columnwidth}{@{\extracolsep{\fill}}lcccc@{}}
\toprule
 & \multicolumn{2}{c}{HumanEval} & \multicolumn{2}{c}{MBPP} \\
\cmidrule(lr){2-3}\cmidrule(lr){4-5}
Method & PFLOPs/query & Time/query (s) & PFLOPs/query & Time/query (s) \\
\midrule
Ours          & \textbf{0.107} & \textbf{1.5} & \textbf{0.145} & \textbf{2.6} \\
MASRouter     & 0.200 & 4.4 & 0.289 & 6.4 \\
Puppeteer-32B & 0.144 & 3.0 & 0.163 & 3.2 \\
\bottomrule
\end{tabular*}
\caption{Approximate inference FLOPs and batch-amortized wall-clock time per query on HumanEval and MBPP.}
\label{tab:hw_cost}
\end{table}

\section{Per-Dataset Accuracy and Cost}
\label{app:cost_detail}

Tables~\ref{tab:gsm_cost} and~\ref{tab:he_cost} report per-method accuracy and
total test-set cost on GSM-Hard and HumanEval, the data underlying the Pareto
comparison of Figure~\ref{fig:pareto}. Cost is the size-based reference cost of
Appendix~\ref{app:cost}. Figure~\ref{fig:pareto_gsm} shows the GSM-Hard
accuracy--cost frontier, the companion to the HumanEval frontier in the main
text.

\begin{figure}[h]
    \centering
    \includegraphics[width=\columnwidth]{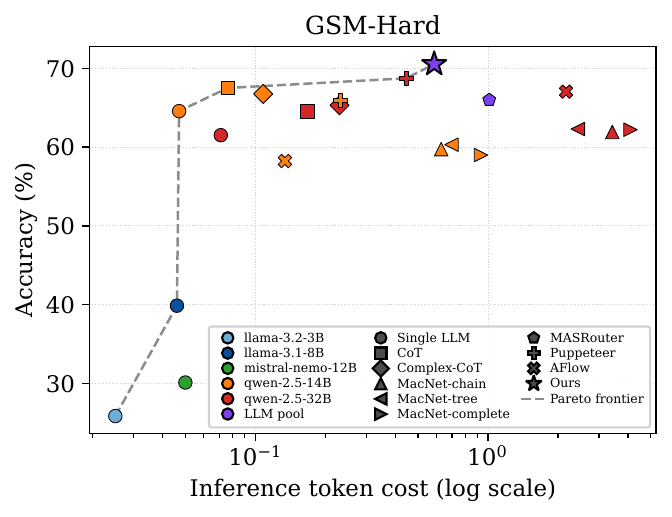}
    \caption{Accuracy versus inference token cost on GSM-Hard, with cost on a log scale. As on HumanEval, our method lies on the high-accuracy end of the Pareto frontier, reaching the highest accuracy among compared methods at substantially lower cost than the strongest multi-agent and routing baselines.}
    \label{fig:pareto_gsm}
\end{figure}

\begin{table}[h]
\centering
\small
\setlength{\tabcolsep}{5pt}
\begin{tabular}{llcc}
\toprule
Method & LLM & Acc.\ (\%) & Cost \\
\midrule
\multirow{5}{*}{Single LLM}
& llama-3.2-3B     & 25.85 & 0.025 \\
& llama-3.1-8B     & 39.87 & 0.046 \\
& mistral-nemo-12B & 30.11 & 0.050 \\
& qwen-2.5-14B     & 64.58 & 0.047 \\
& qwen-2.5-32B     & 61.52 & 0.071 \\
\midrule
\multirow{2}{*}{CoT}         & qwen-2.5-14B & 67.52 & 0.076 \\
                             & qwen-2.5-32B & 64.56 & 0.167 \\
\multirow{2}{*}{Complex-CoT} & qwen-2.5-14B & 66.76 & 0.108 \\
                             & qwen-2.5-32B & 65.32 & 0.230 \\
\midrule
\multirow{2}{*}{MacNet-Chain}    & qwen-2.5-14B & 59.75 & 0.629 \\
                                 & qwen-2.5-32B & 61.93 & 3.421 \\
\multirow{2}{*}{MacNet-Tree}     & qwen-2.5-14B & 60.32 & 0.696 \\
                                 & qwen-2.5-32B & 62.31 & 2.433 \\
\multirow{2}{*}{MacNet-Complete} & qwen-2.5-14B & 59.00 & 0.933 \\
                                 & qwen-2.5-32B & 62.22 & 4.099 \\
\midrule
\multirow{2}{*}{AFlow}     & qwen-2.5-14B & 58.24 & 0.134 \\
                           & qwen-2.5-32B & 67.05 & 2.165 \\
\multirow{2}{*}{Puppeteer} & qwen-2.5-14B & 65.91 & 0.232 \\
                           & qwen-2.5-32B & 68.75 & 0.445 \\
MASRouter & LLM Pool & 66.00 & 1.012 \\
\midrule
Ours & LLM Pool & 70.55 & 0.587 \\
\bottomrule
\end{tabular}
\caption{Accuracy and total test-set cost on GSM-Hard.}
\label{tab:gsm_cost}
\end{table}

\begin{table}[h]
\centering
\small
\setlength{\tabcolsep}{5pt}
\begin{tabular}{llcc}
\toprule
Method & LLM & Acc.\ (\%) & Cost \\
\midrule
\multirow{5}{*}{Single LLM}
& llama-3.2-3B     & 62.79 & 0.001 \\
& llama-3.1-8B     & 69.78 & 0.0025 \\
& mistral-nemo-12B & 68.22 & 0.002 \\
& qwen-2.5-14B     & 82.95 & 0.003 \\
& qwen-2.5-32B     & 84.37 & 0.009 \\
\midrule
\multirow{2}{*}{CoT}         & qwen-2.5-14B & 82.17 & 0.004 \\
                             & qwen-2.5-32B & 85.15 & 0.012 \\
\multirow{2}{*}{Complex-CoT} & qwen-2.5-14B & 80.62 & 0.005 \\
                             & qwen-2.5-32B & 85.47 & 0.014 \\
\midrule
\multirow{2}{*}{MacNet-Chain}    & qwen-2.5-14B & 86.82 & 0.019 \\
                                 & qwen-2.5-32B & 83.80 & 0.059 \\
\multirow{2}{*}{MacNet-Tree}     & qwen-2.5-14B & 82.95 & 0.024 \\
                                 & qwen-2.5-32B & 86.02 & 0.084 \\
\multirow{2}{*}{MacNet-Complete} & qwen-2.5-14B & 84.50 & 0.033 \\
                                 & qwen-2.5-32B & 86.05 & 0.119 \\
\midrule
\multirow{2}{*}{AFlow}     & qwen-2.5-14B & 85.27 & 0.014 \\
                           & qwen-2.5-32B & 84.50 & 0.224 \\
\multirow{2}{*}{Puppeteer} & qwen-2.5-14B & 83.59 & 0.029 \\
                           & qwen-2.5-32B & 85.16 & 0.047 \\
MASRouter & LLM Pool & 85.16 & 0.057 \\
\midrule
Ours & LLM Pool & 89.84 & 0.032 \\
\bottomrule
\end{tabular}
\caption{Accuracy and total test-set cost on HumanEval.}
\label{tab:he_cost}
\end{table}

\section{Variance across Training Seeds}
\label{app:seed_variance}

Table~\ref{tab:main_results} reports one training run per method. To quantify
run-to-run variability, we retrained our method and the two strongest baselines
with three independent seeds each and report the per-benchmark mean and sample
standard deviation in Table~\ref{tab:seed_variance}, together with the overall
mean (the macro-average over benchmarks for each seed, then mean and standard
deviation across seeds). Our method has the highest mean on all five benchmarks
and improves the overall mean by $4.91$ points over MASRouter and $2.10$ over
Puppeteer-32B. The ranking is identical under every seed: the lowest overall
score among our seeds ($76.62$) exceeds the highest among Puppeteer's ($75.79$)
and MASRouter's ($74.27$).

\begin{table*}[h]
\centering
\small
\setlength{\tabcolsep}{4pt}
\begin{tabular}{lcccccc}
\toprule
Method & GSM-Hard & MATH & MBPP & HumanEval & MMLU-Pro & Overall \\
\midrule
MASRouter     & 65.37 $\pm$ 0.55 & 74.37 $\pm$ 2.28 & 76.87 $\pm$ 2.21 & 81.51 $\pm$ 3.94 & 64.09 $\pm$ 4.83 & 72.44 $\pm$ 1.88 \\
Puppeteer-32B & 66.98 $\pm$ 1.87 & 77.17 $\pm$ 4.41 & 76.41 $\pm$ 2.29 & 86.72 $\pm$ 2.06 & 68.98 $\pm$ 1.01 & 75.25 $\pm$ 0.52 \\
Ours          & \textbf{69.76 $\pm$ 1.45} & \textbf{78.37 $\pm$ 1.88} & \textbf{80.40 $\pm$ 1.21} & \textbf{88.54 $\pm$ 1.19} & \textbf{69.62 $\pm$ 1.17} & \textbf{77.35 $\pm$ 0.63} \\
\bottomrule
\end{tabular}
\caption{Mean $\pm$ sample standard deviation of test accuracy (\%) over three training seeds per method. Overall is the macro-average over the five benchmarks.}
\label{tab:seed_variance}
\end{table*}

\section{Gate Behavior at Test Time}
\label{app:gate_behavior}

Table~\ref{tab:gate_behavior} checks that the trained gates do not collapse to
trivial policies. The write rate is the fraction of reasoning steps committed to
memory; the retrieved fraction is the mean number of retrieved items divided by
the number available at that step (the memory is empty at step~0). Write rates
lie strictly inside $(0,1)$ on all three benchmarks, and the retrieval gate becomes
more selective as memory grows rather than exposing all or none of it. The
per-step entropy of the gate distributions also stays well above zero throughout
training on all three benchmarks.

\begin{table}[h]
\centering
\scriptsize
\setlength{\tabcolsep}{3pt}
\renewcommand{\arraystretch}{1.05}
\begin{tabular*}{\columnwidth}{@{\extracolsep{\fill}}lccl@{}}
\toprule
Benchmark & Mean depth & Write rate & Retrieved fraction, steps 1--5 \\
\midrule
GSM-Hard  & 2.92 & 39.6\% & 16 / 24 / 14 / 12 / 11\% \\
MBPP      & 3.79 & 32.8\% & 9 / 30 / 20 / 15 / 12\% \\
HumanEval & 3.30 & 36.5\% & 91 / 48 / 33 / 24 / 21\% \\
\bottomrule
\end{tabular*}
\caption{Test-time gate behavior of the full system: mean realized depth, fraction of steps written to memory, and fraction of available memory items retrieved at each step.}
\label{tab:gate_behavior}
\end{table}

\section{Halting and Training-Time Computation}
\label{app:halting_training}

Adaptive halting also reduces training-time computation, since shorter
trajectories mean fewer rollout steps per update.
Table~\ref{tab:halting_training} compares the mean realized depth and the total
number of rollout steps over training with and without halting.

\begin{table}[h]
\centering
\scriptsize
\setlength{\tabcolsep}{3pt}
\renewcommand{\arraystretch}{1.05}
\begin{tabular*}{\columnwidth}{@{\extracolsep{\fill}}lccl@{}}
\toprule
Benchmark & Depth, halting & Depth, no halting & Rollout steps, halting vs.\ none \\
\midrule
GSM-Hard  & 3.4 & 6.0 & 26.5k vs.\ 47.9k ($1.81\times$) \\
HumanEval & 3.3 & 6.0 & 6.3k vs.\ 11.5k ($1.83\times$) \\
\bottomrule
\end{tabular*}
\caption{Effect of adaptive halting on training-time computation.}
\label{tab:halting_training}
\end{table}

\section{Agent Role Profiles}
\label{app:roles}

The role allocator selects among the same $26$ heterogeneous role profiles used by MASRouter~\cite{yue2025masrouter}, grouped into three domains: $7$ code-oriented roles, $7$ commonsense/knowledge roles, and $12$ mathematical-reasoning roles. Each profile is a short natural-language description that we encode and project into the latent role space (Section~\ref{sec:role_selector}). At instantiation, this description, together with the role's reasoning strategy and the benchmark's output-format requirements, forms the agent's system prompt, while the user prompt supplies the query and any retrieved context. The descriptions are listed below.
The imbalance across domains reflects the original MASRouter catalog, which we
adopt unchanged so that both methods select from the same roles. To test
dependence on the full catalog, we randomly subsampled it to $13$ roles, keeping
roughly half of each domain, and retrained the full system: accuracy is $70.74$
on GSM-Hard and $89.06$ on HumanEval, against $70.55$ and $89.84$ with all $26$
roles, so performance does not hinge on the full role set.

\paragraph{Code.}
\begin{itemize}\setlength{\itemsep}{2pt}
\item \textbf{AlgorithmDesigner.} Specifies the design of the algorithm, including explanations, usage instructions, and API references, optionally giving pseudocode for the main logic; replies concisely.
\item \textbf{ProgrammingExpert.} A programming expert who, given a function signature and docstring, writes the full implementation (restating the signature) in a single Python code block.
\item \textbf{BugFixer.} A programming expert who restates the signature and returns a corrected full implementation in a Python code block.
\item \textbf{ReflectProgrammer.} A programming expert who reflects on prior attempts and returns a full implementation in a Python code block.
\item \textbf{PlanSolver.} Produces the pseudocode of the target function.
\item \textbf{ProjectManager.} Oversees the overall code structure, suggests optimal design patterns for maintainability and flexibility, and avoids over-engineering; replies concisely.
\item \textbf{TestAnalyst.} Identifies problems in the current code from test data and feedback, supplies special cases and boundary conditions to watch, and points out potential errors; replies concisely.
\end{itemize}

\paragraph{Commonsense and Knowledge.}
\begin{itemize}\setlength{\itemsep}{2pt}
\item \textbf{KnowledgeExpert.} A knowledgeable question-answering expert who analyzes step by step and selects the correct answer.
\item \textbf{Reflector.} Re-examines the question-answering process step by step and selects the correct answer.
\item \textbf{Critic.} Points out potential issues in other agents' analyses point by point and gives a critical opinion before the final result.
\item \textbf{Scientist.} A scientist with natural-science knowledge who provides a thorough solving process, including necessary proofs and explanations.
\item \textbf{Economist.} An experienced economist (macroeconomics, microeconomics, financial markets) who gives well-reasoned, evidence-based answers.
\item \textbf{Historian.} Analyzes cultural, economic, political, and social events from primary sources to reason about the past.
\item \textbf{WikiSearcher.} Lists the key Wikipedia entities that should be looked up to solve the problem.
\end{itemize}

\paragraph{Mathematical Reasoning.}
\begin{itemize}\setlength{\itemsep}{2pt}
\item \textbf{MathSolver.} A math expert who produces a solving process from the hints supplied by other agents.
\item \textbf{Mathematician.} A mathematician skilled at math games, arithmetic, and long-horizon planning.
\item \textbf{MathTeacher.} Teaches the solution step by step as if to a student.
\item \textbf{MathAnalyst.} First derives the solution symbolically (variables as letters), then substitutes values to compute the result.
\item \textbf{Inspector.} Checks whether the problem-solving logic, calculations, and any accompanying code are correct and consistent, then gives its own step-by-step solution.
\item \textbf{AlgorithmEngineer.} Integrates step-by-step reasoning with Python code to solve the problem.
\item \textbf{ProgrammingExpert.} Analyzes the problem and writes functions, combining reasoning with Python code.
\item \textbf{SoftwareDeveloper.} Designs efficient solutions and provides clear, concise functions.
\item \textbf{Engineer.} An experienced engineer who solves the problem from engineering knowledge.
\item \textbf{Scientist.} Provides a detailed solving process with necessary proofs and explanations.
\item \textbf{Economist.} Applies economic reasoning to give evidence-based answers.
\item \textbf{CertifiedAccountant.} Analyzes financial problems and returns correct calculations and solutions.
\end{itemize}

\section{Aggregator Prompt}
\label{app:aggregator_prompt}

Once the halting controller stops the trajectory, the Aggregator LLM receives the query and the retained execution records and produces the final answer. The aggregator is instantiated with the backbone the router selected most frequently during the trajectory (Section~\ref{sec:implementation_details}). Across benchmarks the aggregator shares a common framing in its system prompt, namely to weigh the analyses and results of the other agents, identify errors, and commit to a single most-reliable answer, while the user prompt enforces the benchmark-specific output format. The boxes below paraphrase the system and user prompts for each benchmark.

\begin{promptbox}{Aggregator prompt: MATH}
\textbf{System.} Weigh the other agents' analyses and code, justify the decision, and report the final numeric answer as a boxed value with no units (e.g., ``The answer is $\boxed{140}$'').\\[2pt]
\textbf{User.} Give the final answer in the form ``The answer is $\boxed{140}$''.
\end{promptbox}

\begin{promptbox}{Aggregator prompt: GSM-Hard}
\textbf{System.} Weigh the other agents' analyses and code and commit to the most reliable answer, with reasons. Report a single pure number, with no units, symbols, commas, or scientific notation, and no rounding unless the problem asks; the last line must read exactly \texttt{The answer is <number>}.\\[2pt]
\textbf{User.} Give the final answer with a last line of the form \texttt{The answer is <number>}.
\end{promptbox}

\begin{promptbox}{Aggregator prompt: HumanEval and MBPP}
\textbf{System.} Act as the decision-maker over the other agents' outputs, identify errors, and return the answer as a single Python code block containing nothing else.\\[2pt]
\textbf{User.} Given the function signature and docstring, with any prior designs or implementations, write the full implementation (restating the signature) in one Python code block.
\end{promptbox}

\begin{promptbox}{Aggregator prompt: MMLU-Pro}
\textbf{System.} Act as the decision-maker over the other agents' answers and analyses and identify errors.\\[2pt]
\textbf{User.} Exactly one of up to ten options (A to J) is correct; end with a last line of the form ``The answer is $\boxed{X}$'', where X is one of A to J.
\end{promptbox}

\section{Backbone LLM Profiles}
\label{app:llm_profiles}

The LLM router selects among the five open-weight backbones through their natural-language capability descriptions, which a variational encoder maps into the latent backbone space (Section~\ref{sec:llm_router}). The descriptions below are the text encoded for each model; each description also states the model's reference input/output price (Table~\ref{tab:backbones}), so the encoded text covers both capability and cost.

\begin{itemize}\setlength{\itemsep}{2pt}
\item \textbf{llama-3.2-3B.} Meta's compact 3-billion-parameter, text-only instruction-tuned model with a 128k-token context window. The cheapest and fastest choice in the pool, with solid general reasoning and instruction following; suited to easy queries where quality is less critical.
\item \textbf{llama-3.1-8B.} Meta's widely used 8-billion-parameter instruction-tuned model, offering solid general-purpose reasoning and instruction following at very low cost; a reliable baseline across a broad range of tasks.
\item \textbf{mistral-nemo-12B.} Mistral AI's 12-billion-parameter dense model built in partnership with NVIDIA, with strong multilingual performance, a 128k-token context window, and solid general reasoning and coding at modest cost; a reliable mid-size workhorse.
\item \textbf{qwen-2.5-14B.} Alibaba Cloud's 14-billion-parameter dense instruction-tuned model with a 128k-token context window; strong on mathematics, coding, and general reasoning, punching above its size class as a capable mid-tier option for moderately hard queries.
\item \textbf{qwen-2.5-32B.} Alibaba Cloud's flagship dense model in this pool, with frontier-level performance among open-weight models in its size class on mathematics, coding, and complex reasoning; reserved for the hardest queries where smaller models fall short.
\end{itemize}

\end{document}